\documentclass[sn-mathphys-num]{sn-jnl}

\usepackage{graphicx}%
\usepackage{multirow}%
\usepackage{amsmath,amssymb,amsfonts}%
\usepackage{amsthm}%
\usepackage{mathrsfs}%
\usepackage[title]{appendix}%
\usepackage{xcolor}%
\usepackage{textcomp}%
\usepackage{manyfoot}%
\usepackage{booktabs}%
\usepackage{algorithm}%
\usepackage{algorithmicx}%
\usepackage{algpseudocode}%
\usepackage{listings}%

\begin{document}

\title[A Multilateral Attention-enhanced Deep Neural Network for Disease Outbreak Forecasting: A Case Study on COVID-19]{A Multilateral Attention-enhanced Deep Neural Network for Disease Outbreak Forecasting: A Case Study on COVID-19}


\author[1]{\fnm{Ashutosh} \sur{Anshul}}\email{ashutoshanshul01@gmail.com}

\author[1]{\fnm{Jhalak} \sur{Gupta}}\email{gupta.jhalak00@gmail.com}

\author[1]{\fnm{Mohammad Zia} \sur{Ur Rehman}}\email{phd2101201005@iiti.ac.in}

\author*[1]{\fnm{Nagendra} \sur{Kumar}}\email{nagendra@iiti.ac.in}

\affil*[1]{\orgdiv{Computer Science and Engineering}, \orgname{Indian Institute of Technology Indore}, \orgaddress{\street{Simrol}, \city{Indore}, \postcode{453552}, \state{Madhya Pradesh}, \country{India}}}


\abstract{The worldwide impact of the recent COVID-19 pandemic has been substantial, necessitating the development of accurate forecasting models to predict the spread and course of a pandemic. Previous methods for outbreak forecasting have faced limitations by not utilizing multiple sources of input and yielding suboptimal performance due to the limited availability of data. In this study, we propose a novel approach to address the challenges of infectious disease forecasting. We introduce a Multilateral Attention-enhanced GRU model that leverages information from multiple sources, thus enabling a comprehensive analysis of factors influencing the spread of a pandemic. By incorporating attention mechanisms within a GRU framework, our model can effectively capture complex relationships and temporal dependencies in the data, leading to improved forecasting performance. Further, we have curated a well-structured multi-source dataset for the recent COVID-19 pandemic that the research community can utilize as a great resource to conduct experiments and analysis on time-series forecasting. We evaluated the proposed model on our COVID-19 dataset and reported the output in terms of RMSE and MAE. The experimental results provide evidence that our proposed model surpasses existing techniques in terms of performance. We also performed performance gain and qualitative analysis on our dataset to evaluate the impact of the attention mechanism and show that the proposed model closely follows the trajectory of the pandemic.}




\maketitle

\section{Introduction}
Infectious disease outbreaks have a significant worldwide impact, extending beyond national borders and impacting not just people's health and well-being but also the social, political, and economic fabric of entire countries. Historical accounts reveal the far-reaching consequences of pandemics, including the Black Death of the 14th century, the Spanish flu of 1918, and the recent COVID-19 pandemic. Such events disrupt daily life, challenge healthcare infrastructures, and engender substantial economic repercussions. 

For instance, the swift and widespread transmission of COVID-19 across the globe within a remarkably condensed timeframe had a monumental and far-reaching impact on countless nations. Based on data provided by the Indian government as of 6 December 2022, India ranks second globally in terms of confirmed COVID-19 cases, following the United States of America, with a reported total of 4,46,74,842 infections. Additionally, India stands third in terms of COVID-19-related fatalities, trailing behind the United States and Brazil, with 5,30,630 recorded deaths. This outbreak effectively brought a halt to a wide array of activities taking place worldwide, generating an immense sense of apprehension and distress among individuals among people globally.

These infectious diseases have far-reaching effects beyond health, causing disruptions in economies, healthcare systems, education, and social dynamics. Governments worldwide implement measures like lockdowns and travel restrictions to curb the virus spread and ease pressure on healthcare infrastructure. The epidemics lead to significant loss of lives, inadequate healthcare resources, job losses, and economic hardships. Therefore, anticipating the future trend of an epidemic or a pandemic can assist relevant departments and personnel in developing preparedness strategies and implementing policies beforehand to control the spread and outbreak of such a disease which is crucial for the well-being of humankind. This will also enable the neighboring countries which may also be affected by it to allocate resources and plan their own response strategies effectively. 

Harnessing the growing power of artificial intelligence for disease prediction holds tremendous promise in our quest for proactive pandemic preparedness. By leveraging advanced computational techniques, we can delve deeper into understanding the intricate patterns and dynamics of infectious diseases, ultimately aiding in developing effective strategies for prevention, early detection, and timely response. This will help in overcoming the challenges posed by any infectious disease and build a more resilient and prepared global community by enhancing our ability to detect emerging threats, optimize resource allocation, and implement targeted interventions. 

Several methods have been proposed for forecasting infectious disease outbreaks, such as the recent COVID-19. Castillo \MakeLowercase{\textit{et al.}}~\cite{ref12} suggested a fusion method for COVID-19 time series forecasting that included fractal theory and fuzzy logic. Ma \MakeLowercase{\textit{et al.}}~\cite{ref34} proposed using a Markov model and Long Short-Term Memory (LSTM) combination to forecast COVID-19 instances worldwide. Bedi \MakeLowercase{\textit{et al.}}~\cite{ref2} suggested using both LSTM and a modified susceptible-exposed-infected-recovered-deceased (SEIRD) model to forecast COVID-19 cases in India, and they evaluated how well they performed against one another. However,  the currently proposed models only use daily statistics of the outbreak, such as the number of cases, deaths, and recoveries in the region of interest for building their forecasting model, which might not give an accurate picture of the data as other important factors that could affect the outbreak may exist. Secondly, the limited size of available data can impact the model's performance. Moreover, these models may fail to understand any interdependencies present between data coming from multiple sources.

Unlike most of the existing works, our model utilizes data from multiple sources including News, Google trends, and outbreak-related statistics to effectively forecast the count of the cases. In addition to the number of cases and deaths, we also extracted other related data that played some role in the spread of the outbreak and the number of new cases, such as vaccination, number of tests, old age population, hospital-related data, cases in neighboring regions or regions involved in the exchange of population, etc. The inclusion of these additional factors expanded the dataset and captured different aspects and perspectives that helped in forecasting cases with a more holistic view. 

The proposed model implements a multi-head attention block that accepts the data from multiple sources, understands the underlying dependencies, and generates a rich intermediate representation. By employing the attention mechanism coupled with GRU-based forecasting, our approach was able to capitalize on the complementary strengths of each model while minimizing their individual weaknesses. We evaluated our model on the COVID-19 prediction use case for India. We observed improved prediction accuracy, robustness, and generalization capabilities, ultimately enhancing the overall quality of our predictions. We implemented other existing models and compared their performance with our proposed model. Our model surpassed the efficacy of other existing methods. Lastly, we performed experiments to study the contribution of each of the features and the impact of the attention mechanism in improving our model's accuracy. The key contributions of this study are outlined as follows:
\begin{enumerate}

\item We propose a novel deep-learning framework to address the problem of forecasting the number of cases of an infectious disease outbreak. We extract important factors influencing the progression of the pandemic with the aim of precisely capturing its dynamics. We also perform further analysis which shows each of these factors has caused significant improvement in the model's performance.
\item We propose a deep-learning-based Multilateral Attention-enhanced GRU (MAG) model, which is a combination of multi-head attention followed by Gated Recurrent Unit (GRU) layers. The model is thus able to understand the interdependencies present in the dataset collected from multiple sources and then pass the attention block's output to GRU layers for temporal prediction. The model can also be utilized to address other temporal forecasting problems.
\item  We construct a well-structured COVID-19 dataset by extracting (i) News headlines related to COVID-19, (ii) Google trends data for COVID-19-related keywords, and (iii) the daily number of COVID-19 tests, positive cases, and deaths due to COVID-19, etc. This comprehensive dataset serves as the basis for evaluating our model's performance and conducting additional analyses.
\item We evaluate the proposed model against other existing methods. Our model outperforms the existing methods by incorporating data from diverse and relevant sources and employing the attention mechanism. We analyze the impact of each data source and the attention block on the model's efficacy and present significant qualitative findings based on the COVID-19 dataset.

\end{enumerate}

The subsequent sections of this paper are structured as follows: Section \ref{sec:sa_rw} provides an overview of the research related to our work. Section \ref{sec:sa_meth} gives a comprehensive explanation of our proposed methodology. Section \ref{secExperimental} encompasses details about the dataset and a description of the conducted experiments. It also discusses the results obtained along with analysis and comparisons. Lastly, Section \ref{sec:sa_cnfw} concludes the work.

\section{Related Works}
\label{sec:sa_rw}
In this section, we present a comprehensive review of prior works pertaining to time series forecasting. Firstly, we discuss some relevant works centered on rule-based forecasting techniques. Following that, in the two subsequent subsections, we explore the methodologies employed using machine learning and deep learning for forecasting purposes. Finally, we provide an analysis of some of the previous works specifically focusing on the forecasting of COVID-19 case numbers.

\subsection{Rule-based Time Series Forecasting}
Rule-based time series forecasting incorporates specific rules or guidelines based on the understanding of the underlying patterns, behaviors, or characteristics of the time series data. These rules can be derived from domain expertise, historical observations, or empirical relationships and maybe logical conditions, thresholds, or predefined mathematical formulas that guide the forecasting process.

Several rule-based methods ~\cite{ref12, ref5, ref6, ref7, ref8, ref9, ref10, ref11} have been proposed for addressing time series forecasting tasks. Arora \MakeLowercase{\textit{et al.}} ~\cite{ref5} proposed rule-based modification to triple seasonal autoregressive moving average (SARMA) ~\cite{ref13}, triple seasonal Holt-Winters-Taylor (HWT) exponential smoothing ~\cite{ref13}, and triple seasonal intra-week singular value decomposition (SVD)-based exponential smoothing and implemented artificial neural networks (ANNs) to predict the load specifically on anomalous days, like public days. In their later work, Arora \MakeLowercase{\textit{et al.}} ~\cite{ref10} also used the rule-based SARMA model for predicting load in France. They made adjustments to the basic rule-based SARMA model to enable it to adapt to the France dataset and for special days. For multidimensional time series forecasting, Borisov \MakeLowercase{\textit{et al.}} ~\cite{ref7} addressed the challenge of understanding the interdependency of each time series and simplifying the task to individual one-dimensional time series. It proposed structural and parametrical modifications of fuzzy rule-based cognitive models to enhance forecasting accuracy by incorporating direct and indirect interdependencies. Hakimi \MakeLowercase{\textit{et al.}} ~\cite{ref9} implemented a novel reward-based time-series forecasting model (RBTM) for time series forecasting. It extracts features as rules and calculates their rewards. Further, it uses the rules, rewards, and the prior time series for predicting the next value.

While rule-based forecasting can provide insights, interpretability, and explainability in the forecasting process by leveraging expert knowledge, it heavily relies on the accuracy and relevance of the predefined rules, which may not always capture all the complex patterns or adapt well to changing dynamics in the time series. Therefore, the effectiveness of rule-based forecasting depends on the quality of the rules and their alignment with the characteristics of the data being forecasted.

\subsection{Machine Learning-based Time Series Forecasting}
Time series forecasting has made extensive use of machine learning techniques like Autoregressive Integrated Moving Average (ARIMA) and Seasonal ARIMA. Linear time series forecasting frequently uses ARIMA models. Some statistical analysis and data prediction addressed using ARIMA include predicting the price of electricity ~\cite{ref20}, the demand for energy ~\cite{ref21}, the velocity of a vehicle ~\cite{ref22}, future inflation rate~\cite{ref18}, and stock indices ~\cite{ref23}. However, the conventional ARIMA model is inappropriate for nonlinear time series. Hence several complex models have been proposed to enhance the performance of ARIMA on real-life time series data. Li \MakeLowercase{\textit{et al.}} ~\cite{ref24} proposed a complex neuro-fuzzy-based ARIMA model for financial time series forecasting. Ordóñez \MakeLowercase{\textit{et al.}} ~\cite{ref25} used an ARIMA-SVM hybrid model for predicting the remaining useful life of an aircraft. Sharma \MakeLowercase{\textit{et al.}} ~\cite{ref15} proposed recursively using eigenvalue decomposition of Hankel matrix (EVDHM) over the input data to break it down into subcomponents and reducing the non-stationarity. The final output of the decomposition is then fed to ARIMA for training and prediction.

SARIMA, which stands for Seasonal ARIMA, is a variant of the ARIMA that incorporates seasonal components to handle time series with seasonal patterns. It takes into account seasonal differences and autocorrelations to provide more accurate forecasts. It has been used in multiple analyses and prediction tasks, for example, crop yield prediction~\cite{ref26}, traffic accidents ~\cite{ref27}, nifty-500 stock market prediction ~\cite{ref16}, etc.

However, traditional machine-learning models like ARIMA and SARIMA are likely to struggle to capture complex non-linear patterns or relationships present in time series data. They often require manual feature engineering, where domain expertise is necessary to select and construct pertinent features from the raw input data. Further, traditional machine-learning models may also struggle to capture long-term dependencies in time series data. Basic deep learning models like LSTM overcome these shortcomings and hence significantly outperform ARIMA and SARIMA in time series prediction~\cite{ref14, ref17}.

\subsection{Deep Learning-based Time Series Forecasting}
Deep learning models like GRU, LSTM, Bidirectional GRU (BiGRU), and Bidirectional LSTM (BiLSTM) are among the most popular and emerging approaches for addressing time series forecasting. Several methods have used LSTMs for their time series prediction. Ghanbari \MakeLowercase{\textit{et al.}} ~\cite{ref28} to analyze household electricity consumption data and predict the electricity consumption in the next seven days. Wang \MakeLowercase{\textit{et al.}} ~\cite{ref29} analyzed and compared the LSTM performance for multi-step time series prediction using two methods, multi-step input, and seq2vec.

Zhai \MakeLowercase{\textit{et al.}} ~\cite{ref30} proposed a combination of the Gated Recurrent Unit (GRU) and eXtreme Gradient Boosting (XGBoost) models to address multivariate time series forecasting. They evaluated the model by predicting the heating furnace's temperature and found that the hybrid XGB-GRU model outperformed the single GRU and XGBoost models. Honjo \MakeLowercase{\textit{et al.}} ~\cite{ref31} proposed pooling attention and GRU-based framework for time series prediction. The model consists of a CNN layer for feature extraction, two attention mechanisms for capturing the underlying patterns in data, and a GRU layer for final forecasting. Xie \MakeLowercase{\textit{et al.}} ~\cite{ref32} implemented a GRU-based model for predicting melt spinning properties. Their GRU model surpassed the LSTM and other models over their data. Wang \MakeLowercase{\textit{et al.}} ~\cite{ref33} implemented ARIMA, LSTM, BiLSTM, and BiGRU for traffic flow prediction and found BiGRU to be of superior performance.

\subsection{COVID-19 Cases Prediction using Time Series Analysis}
Several methods ~\cite{ref1, ref2, ref3, ref12, ref19, ref34, ref35, ref36} were proposed for addressing the specific problem of predicting the COVID-19 case count across the world. Castillo \MakeLowercase{\textit{et al.}}~\cite{ref12} proposed a fusion approach that combined fuzzy logic and fractal theory for forecasting the COVID-19 time series. The fuzzy logic represents the uncertainty in making the forecast, while the fractal theory is used to evaluate the intricate nature of the data dynamics. Jain \MakeLowercase{\textit{et al.}}~\cite{ref19} employed the ARIMA and SARIMA models to forecast the Covid-19's trajectory. They also used the SARIMAX model to analyze the impact of festive seasons on the pandemic's trajectory.

Ma \MakeLowercase{\textit{et al.}}~\cite{ref34} used a combination of the Markov model and LSTM for predicting COVID-19 cases across the globe. While the LSTM was employed for predicting cases, the Markov model was employed to predict the error between the original data and LSTM's predicted value. The predicted error was used to correct the LSTM's prediction and make the final forecast. Alabdulrazzaq \MakeLowercase{\textit{et al.}}~\cite{ref35} evaluated the ARIMA model for COVID-19 prediction in Kuwait. They performed parameter optimization and used the best-fit model for the forecast. Finally, they evaluated the association between actual and predicted values using Pearson’s correlation coefficient. K\"{u}lah \MakeLowercase{\textit{et al.}}~\cite{ref39} represented the COVID-19 time series data as Gaussian Mixture Models. They proposed a Shifted Gaussian Mixture Model coupled with Similarity-based Estimation to predict a specific country's daily new case values by analyzing similar patterns in other countries.

Shahid \MakeLowercase{\textit{et al.}}~\cite{ref3} employed Support Vector Regression(SVR), ARIMA, LSTM, GRU, and BiLSTM to predict COVID-19 cases, recoveries, and deaths for ten countries. Qu \MakeLowercase{\textit{et al.}}~\cite{ref38} proposed an ensemble of Backpropagation neural network (BPNN), Elman neural network (ENN), Adaptive Neuro-fuzzy Inference System (ANFIS) and LSTM for the task of COVID-19 forecasting. Furthermore, they proposed a heuristic algorithm Sine Cosine Algorithm-Whale Optimization Algorithm (SCOWA) for optimizing the weights of the ensemble. Ali \MakeLowercase{\textit{et al.}}~\cite{ref40} proposed the use of stacked Bi-directional LSTM for accurate forecasting of COVID-19 in South Korea and compared its performance with LSTM and traditional time series models.

All of the above prediction methods analyze the statistical data such as the number of recoveries, cases, and deaths per day. However, very few of them explored the use of other data sources, which provide relevant information about the spread of the pandemic and are highly correlated to the pandemic's trend. In this paper, we explored the use of News and Google trends alongside the daily statistics of COVID-19. We also included other statistical data apart from cases and deaths, for instance, tests, vaccination, age distribution, etc. Finally, to understand the inter-dependencies present in the data from multiple sources, we proposed an attention-enhanced GRU model for predicting the number of cases three days prior.

\section{Proposed Methodology}
\label{sec:sa_meth}
\begin{figure*}[!t]
\centering
\includegraphics[width=5in]{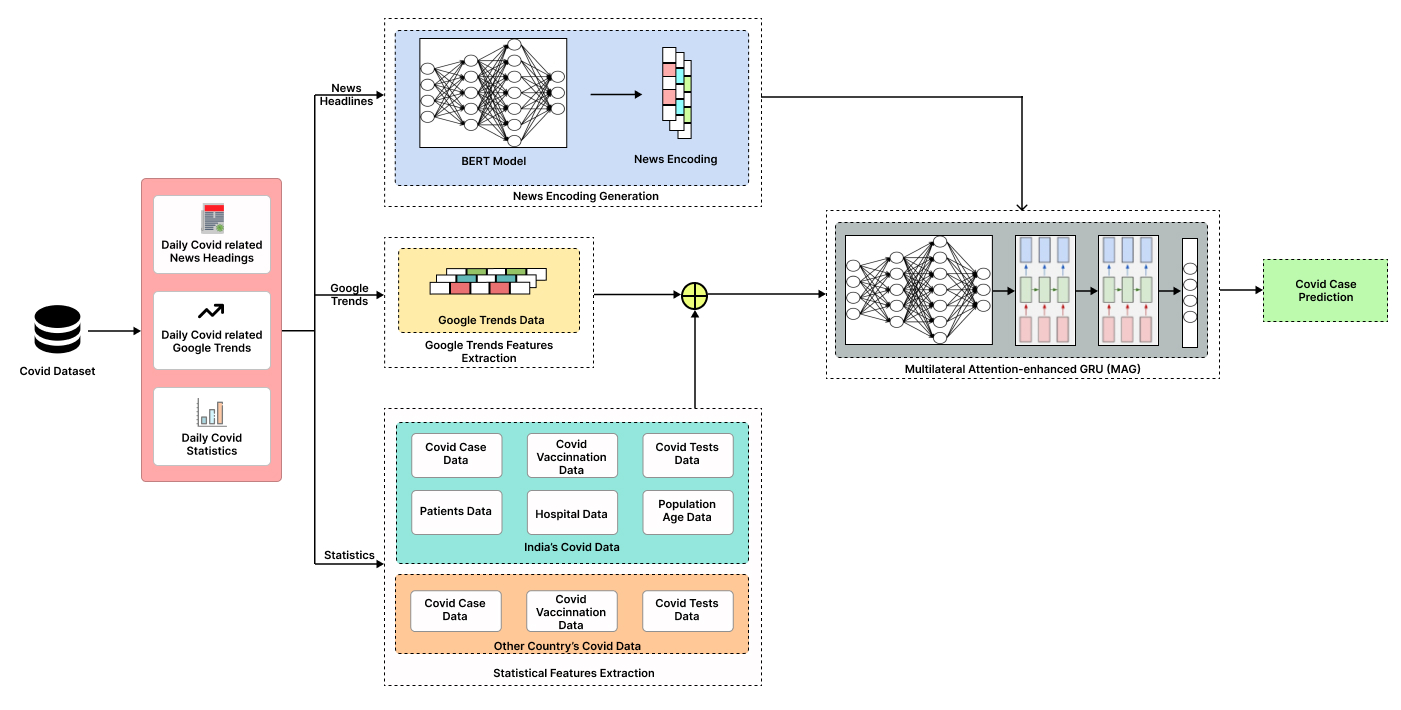}
\caption{System architecture of the proposed model}
\label{sysArch}
\end{figure*}

In this section, we illustrate the architecture of the proposed model as shown in Fig \ref{sysArch}. 
Given the set of news headlines, Google search trends, and daily COVID-19-related statistical data, our model aims to predict the number of new COVID-19 cases in India. The end-to-end system includes three steps (i) extracting news headlines, pandemic-related statistical data of multiple countries, and Google trends data for each day and storing it in a database, (ii) processing the data to generate news embedding, extracting statistical and Google trends features, and (iii) feeding the final feature set into a Multilateral Attention-enhanced GRU (MAG) model for predicting the count of COVID-19 cases 3 days prior. 

\subsection{News Encoding Generation}
\label{secNewsEncodeing}
The outbreak of COVID-19 paused most of the activities going around the world and became the center of attraction for everyone. Hence, there was a significant amount of news articles dedicated to COVID-19 every day. These articles covered a wide range of subjects related to coronaviruses, such as advisories issued by WHO and governments, scientific findings regarding symptoms and the virus, and the daily number of cases and deaths in different parts of the world. The daily amount of increase and decrease in cases of COVID-19 and the symptoms related to COVID-19 were also covered in news headlines. Hence, extracting and processing news related to disease would help in analyzing the daily trend of the pandemic. Since the headlines of the news articles set up a concise context about its content and often include the numerical data related to daily statistics, we extracted just the headlines of all the news articles related to COVID-19 and its symptoms. Once the news headlines were extracted, we converted the raw text into a numerical embedding by passing these headlines through the pre-trained BERT model. 

Bidirectional Encoder Representations from Transformers (BERT) ~\cite{ref4} is a multi-layered transformer encoder-based architecture, which utilizes the bidirectional training capability of transformers for language modeling. A basic transformer consists of an encoder and a decoder, both incorporating attention mechanisms. The encoder processes textual input and produces its corresponding numerical encoding. On the other hand, the decoder utilizes this vector representation to make predictions. Since BERT is a linguistic encoding model to generate text encodings, it adopts multiple layers of the encoder model for training and prediction.

Rather than processing the input series in a traditional right-to-left or left-to-right manner, the BERT model processes the complete input sequence as a whole. Hence, it is considered to be bidirectional, which enables it to comprehend the semantics of a word by analyzing its surrounding context. It is pre-trained on Books Corpus and English Wikipedia through two unsupervised approaches for training, namely, Masked Language Modelling (MLM), and Next Sentence Prediction (NSP). The MLM converts 15\% of the input tokens to [MASK] tokens and then attempts to predict the actual token by understanding the context of surrounding tokens. In NSP, the model reads two sentences and predicts if the second sentence follows the first one.

We employed the pre-trained model to generate vector representations for news headlines. The model takes the headline text as input and produces a 786-dimensional encoding. Finally, we averaged out the encoding of all the headlines for a day to generate our 786-dimensional day-wise news encoding feature.

\subsection{Google Trends Data Extraction}
\label{secGTFeature}
Google Trends tracks the popularity of query terms in Google Search across multiple languages and regions. It has been observed that Google Trends data can be used to analyze various behaviors, such as economic trends, public awareness, and interests in multiple issues. It has even outperformed survey-based data in various forecasting experiments. Hence, extracting Google trends data for terms related to COVID-19 would be helpful in forecasting new COVID cases in India. To achieve this, we extracted Google trends data for 13 terms related to COVID-19 for each day and added it to our dataset. The keywords related to covid used to extract google trends data are `Covid', `Case', `Death', `Fever', `Vaccine', `Wave', `Precaution', `Pandemic', `Delta', `Corona', `Coronavirus', `SARS-CoV-2', `Omicron'.

\subsection{Statistical Feature Extraction}

The statistical data includes various kinds of numerical information related to the COVID-19 outbreak. While the daily count of COVID-19 cases impacted these numerical data, few of these numbers are significant indicators of the forthcoming days' count of COVID-19 cases. Hence, extracting them would help analyze the daily trend of the COVID outbreak. Furthermore, the covid cases in other countries also affect the cases in India. This can be attributed to factors such as location and public inflow. Hence, we extracted everyday COVID-19 cases, tests, and vaccination data for India and 26 other countries. Additionally, we also extracted population, age, and hospital data for India. The statistical data collected is discussed in more detail in Section \ref{datasetandprocessing}.

\subsection{Covid Case Prediction - MAG Model}

\begin{figure*}[!t]
\centering
\includegraphics[width=5in]{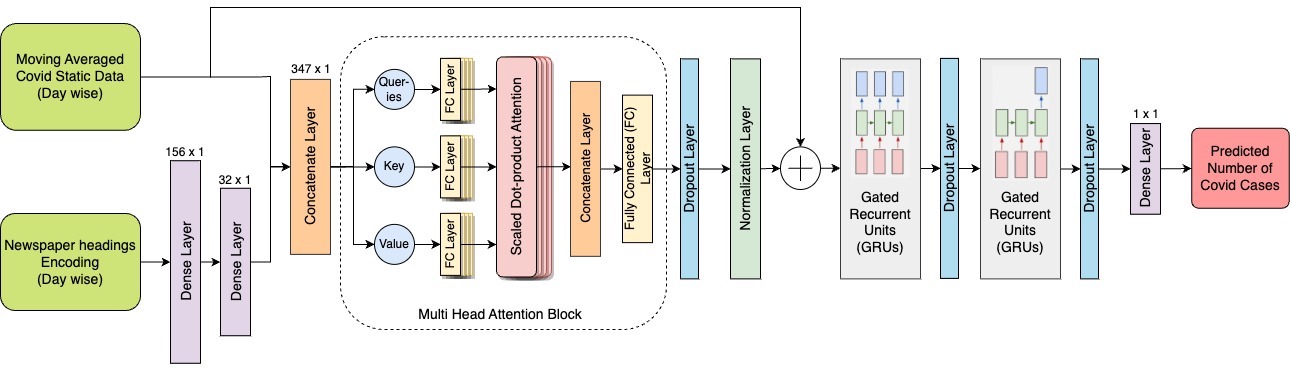}
\caption{Architecture of the MAG model}
\label{atgModel}
\end{figure*}

Once the News encoding, Google Trends data and Statistical features are extracted, the final feature set is passed through our Multilateral Attention-enhanced GRU (MAG) model for predicting the number of new COVID cases. The architecture of the MAG model is illustrated in Fig \ref{atgModel}. It accepts two inputs (i) concatenated feature set for Google Trends and Statistical features, and (ii) News headline encoding. The model first processes the 786-dimensional news encoding through two dense layers of dimensions 156 and 32. The output from the dense layer is concatenated with Google Trends and Statistical feature set, obtained as input. The resultant feature set is then passed to the transformer encoder-based attention layer followed by two GRU layers for the final prediction.

\subsubsection{Transformer Encoder based Attention Layer}
A typical transformer encoder, belonging to seq2seq, is commonly used for feature extraction from sequential data, for instance, a word vector. It takes the sequential vector and its positional encoding as input and passes them through a multi-head attention mechanism followed by a densely connected feed-forward network. This helps it to capture dependency in the data while computing in parallel, thereby learning the semantic information in the input data.

Since our input feature vector consists of information from three different sources, we utilize the multi-head attention block, as shown in Fig \ref{atgModel}, to understand the dependencies existing between the feature sets from different sources and generate a joint representation which is further passed along the model for forecasting. Algorithm \ref{algoMHA} presents the multi-head attention block, which takes three inputs, namely, query, key, and value. In this case, the three inputs are the same, that is, the input feature vector. It then creates $h$ different projections of the query, value, and key by passing them through fully connected layers. Here $h$ is the number of heads taken as an input parameter. On these projections of the query, value, and key, it parallelly performs the Scaled Dot-Product Attention given by Equation \ref{eqScaledProdAtt}, where $Q$ is Query, $V$ is Value, $K$ is Key and $dim_k$ is the dimension of the Key $K$.

\begin{equation}
    \label{eqScaledProdAtt}
    A(Q, V, K) = softmax(\frac{QK^T}{\sqrt{dim_k}})V
\end{equation}

Further, it concatenates the output and passes it through a fully connected layer, resulting in the generation of the output from the multi-head attention block. The output value can be represented by Equation \ref{eqMultiHeadAtt}. Here $W_O, W_i^Q, W_i^V,$ and $W_i^K$ are the weight matrices learned while training.

\begin{equation}
    \label{eqMultiHeadAtt}
        MHA(Q, V, K) = Concat(head_1,..., head_h)W^O
\end{equation}

Here, $head_i = A(QW_i^Q, VW_i^V, KW_i^K)$ and the output from the multi-head attention block is then normalized and added to the original input. The resulting value is then passed to the GRU block for forecasting.

\algnewcommand\algorithmicinput{\textit{Input:}}
\algnewcommand\INPUT{\item[\algorithmicinput]}
\algnewcommand\algorithmicoutput{\textit{Output:}}
\algnewcommand\OUTPUT{\item[\algorithmicoutput]}

\begin{algorithm}
    \caption{Multi-Head Attention}
    \label{algoMHA}
    \begin{algorithmic}[1]
        \INPUT \hspace{8mm} $Q$: Query 
        \Statex \hspace{11.5mm} $V$: Value
        \Statex \hspace{11.5mm} $K$: Key
        \Statex \hspace{11.5mm} $h$: Number Of heads
        \OUTPUT \hspace{4.5mm} $R$: Intermediate Representation Vector
        \Function{Mutli-head Attention}{$Q$, $K$, $V$}
            \For{ $i \gets$ 1 to h in parallel}
                    \State $Q_i \gets QW_i^Q$
                    \State $K_i \gets KW_i^K$
                    \State $V_i \gets VW_i^V$
                    \State $d_i \gets SizeOf(K_i)$
                    \State $F_i \gets \frac{Q_iK_i^T}{\sqrt{d_i}}$
                    \State $A_i \gets softmax(F_i)V_i$
            \EndFor
            \State $R \gets concat(A_1, A_2,....A_h)W^O$
            \State \Return $R$
        \EndFunction
    \end{algorithmic}
\end{algorithm}

\subsubsection{GRU Layer}
Recurrent Neural Network (RNN) is a class of neural networks, which functions using hidden states serving as memory. The output of a hidden state at a given time step is also dependent on its output at the previous timestep. Given an input sequence, $ x = \langle x_1, x_2,....x_3 \rangle$, the hidden state at time step $t$ is defined by equation \ref{eqGRUHiddenState}. Here $W_h$, and $W_x$ are weights, $b$ is the bias and $\sigma$ is the sigmoid function

\begin{equation}
    \label{eqGRUHiddenState}
    h\{t\} = \sigma(W_hh\{t-1\}+W_xx\{t\}+b)
\end{equation}

This allows the RNNs to be well-suited for situations where the input data is of varying length, or the computation requires historical information to be taken into consideration, for instance, speech or text processing, and time-series analysis. However, it becomes difficult to capture long-term dependencies using RNNs, due to vanishing/exploding gradients. This occurs when the time step is significant and the gradients get too small or large. Gated Recurrent Unit (GRU), initially introduced by Cho \MakeLowercase{\textit{et al.}}, is an improved version of RNNs, designed to effectively capture relevant long-term temporal dependencies in sequential data. The GRU achieves this using update and reset gates. The update gate aids the model in determining the extent to which past information should be propagated to the future, while the reset gate enables the model to decide which portions of the past information to disregard. The equations of the update and reset gate at a given time are given by Eq. \ref{eqUpdateGate} and Eq. \ref{eqResetGate} respectively. Here $W_x^u$, $W_h^u$, $W_x^r$, and $W_h^r$ are trainable weights, $x_t$ is the input sequence at timestep $t$, and $h_t$ is the output of the hidden state at timestep $t$. 

\begin{equation}
    \label{eqUpdateGate}
    U_t = \sigma(W_x^ux_t+W_h^uh_{t-1})
\end{equation}

\begin{equation}
    \label{eqResetGate}
    R_t = \sigma(W_x^rx_t+W_h^rh_{t-1})
\end{equation}

Further, the value from the reset gate is used to define memory content, used to store past information. The memory content is represented by Eq. \ref{eqMemoryContent}, where $W_x^c$, $W_h^c$ are trainable weights, $R_t$ is output from reset gate, and $x_t$ and $h_t$ have the same meaning as in Eq. \ref{eqUpdateGate}.

\begin{equation}
    \label{eqMemoryContent}
    C_t = \tanh(W_x^cx_t+R_t \odot W_h^ch_{t-1})
\end{equation}

The transformer encoder layer's output is directed into two GRU layers, each consisting of 100 units. The final GRU layer's output is then forwarded through a dense layer, serving as the output layer, to make predictions regarding the new COVID-19 cases

\section{Experimental Evaluations}
\label{secExperimental}

\subsection{Experimental Setup}
In this section, we discuss the dataset, followed by a discussion on different comparison methods, evaluation metrics used, and hyperparameter tuning.

\subsubsection{Dataset and Preprocessing}
\label{datasetandprocessing}
To evaluate our method, we created our dataset related to COVID-19. We primarily focused on three sources of data, which are, news headlines, COVID-19 related statistical data of multiple countries, and Google trends data related to COVID-19 for each day.

\begin{itemize}
    \item \textbf{News Encoding:} To collect daily news data, we extracted news headlines from The Hindu’s\footnote{https://www.thehindu.com/archive/print/} archive. Furthermore, to filter COVID-19 news, we searched for the presence of the following keywords in the headlines - `covid', `case', `death', `fever', `vaccine', `wave', `precaution', `omicron', `pandemic', `delta', `corona', `coronavirus', `sars-cov-2'. Once the COVID-related headlines are collected for each day, they are processed to generate the encoding to be used for prediction. A detailed discussion on the news encoding generation is presented in section \ref{secNewsEncodeing}.

    \item \textbf{Statistical Data:} To extract the daily statistical data related to COVID-19, we selected the following data and extracted them from Our World in Data repository\footnote{https://github.com/owid/covid-19-data/blob/master/public/data/README.md}.
    \begin{itemize}
        \item COVID Case data - This includes the total and the new number of cases per million population, and the total and new deaths per million population.
        \item COVID Test data - While the number of cases and deaths have direct relation in predicting the upcoming number of cases, tests performed also affect the number of cases. A low test rate is liable to exclude a significant number of COVID positive individuals from the reported count. Hence, we kept a daily track of the total tests performed per thousand people, new tests performed per thousand, and the positive rate of COVID patients,
        \item Vaccination data - The vaccination data includes the total number of vaccination, the total number of vaccinated people, the total number of fully vaccinated people, and the total booster dose administered per hundred individuals.
        \item Hospital-related data - The statistics related to hospitals cover the number of patients admitted to the hospital per million, the number of patients admitted to the ICU per million, and the number of hospital beds present per thousand individuals.
        \item Age-based distribution data - Since elderly people were among the most affected by COVID, we extracted the number of individuals above 60 years of age and the number of individuals above 70 years of age.
        \item Other countries’ statistics - While the above statistics from India directly affect the number of cases in India, it was observed that COVID trends in other countries also affected the trends in India. From the list of countries for which we had the COVID related data available, we manually selected 26 countries. We extracted the COVID cases, COVID tests, and vaccination data for these countries.
    \end{itemize}

    \item \textbf{Google Trends Feature:} A detailed discussion on the google trends features is presented in detail in Section \ref{secGTFeature}.
\end{itemize}

\subsubsection{Comparison Methods}
To assess the performance of the proposed model, we compared it with the following methods:

\begin{itemize}
    \item \textbf{Bedi \MakeLowercase{\textit{et al.}}~\cite{ref2}}: In this work, the authors proposed a modified susceptible–exposed–infected –recovered–deceased (SEIRD) model and LSTM to predict covid-19 trends in India and its four worst-impacted states. They extracted the epidemiological data from covid19india.org website \footnote{https://www.covid19india.org/} between 30th January 2020 to 6th September 2020. Post-training, they predicted the daily recovered, total number of recovered, daily deceased, the total deceased, daily confirmed, and total confirmed numbers for 30 days from 7th September 2020. They also compared the performance of SEIRD and LSTM models and presented a details analysis of the COVID-19 trend in India and its top four worst-impacted states, which are, Karnataka, Maharashtra, Tamil Nadu, and Andhra Pradesh.
    
    \item \textbf{Arora \MakeLowercase{\textit{et al.}}~\cite{ref1}}: The authors trained Stacked LSTM (LSTM), Bi-directional LSTM (Bi-LSTM), and Convolutional LSTM (Conv-LSTM) over 2 months of data of COVID-19 cases in India, ranging from 14th March 2020 to 14th May 2020. They gathered temporal data on covid-19 confirmed cases from the Ministry of Health and Family Welfare \footnote{https://www.mohfw.gov.in/} for each state and union territory in India and aimed to predict the coronavirus cases one day to a week ahead of time. Further, they used the prediction to categorize states and union territories into mild, moderate, and severe zones and discussed the zone-wise prevention measured to control the spread.

    \item \textbf{Shahid \MakeLowercase{\textit{et al.}}~\cite{ref3}}: In this work, Shahid \MakeLowercase{\textit{et al.}} implemented Deep Learning models such as LSTM, Bi-LSTM, and GRU, Machine Learning technique as Support Vector Regression(SVR), and statistical models such as ARIMA to predict COVID-19 confirmed cases, recoveries and deaths for ten countries. They reported the model's performance in terms of  Mean Absolute Error (MAE), Root Mean Square Error (RMSE), and $R^2$ score and presented a comparative analysis of these models.
\end{itemize}

\subsubsection{Evaluation Metric}
We evaluated the efficacy of our proposed model and compared the existing methods in terms of Mean Absolute Error (MAE) and Root Mean Square Error (RMSE). The metrics can be represented by Eq. \ref{eqRMSE}

\begin{equation}
    \label{eqRMSE}
    \begin{split}
        RMSE = \sqrt{\frac{\sum_{i=1}^{N}|pred_i-true_i|^2}{N}} \\
        MAE = \frac{\sum_{i=1}^{N}|pred_i-true_i|}{N}
    \end{split}
\end{equation}

For our use case, $N$ is the number of days, $pred_i$ is the predicted number of COVID-19 cases and $true_i$ is the actual number of COVID-19 cases.

\subsubsection{Hyperparameter Tuning}
For the proposed model, there are six parameters to optimize: (i) batch size ($batch\_size$), (ii) learning rate ($lr$), (iii) number of units in a GRU layer ($gru\_units$), (iv) dropout ($dropout$), (v) number of attention heads in the Multi-head Attention block ($num\_heads$), (iv) size of each attention head in the Multi-head Attention block ($head\_size$). In order to determine the optimal value of each parameter, we implemented the grid search technique and selected the combination of values that produced the least RMSE value. The values used to select $batch\_size$ through grid search were 16, 32, 64, and 128. For $lr$, we used 0.001, 0.01, and 0.1 as possible values to select from. $gru\_units$ were selected between 10, 50, 100, and 150. Similarly, 0.2 and 0.5 were the options codes for $dropout$. $num\_heads$ was selected from 2, 4, 5, and 8. Finally, the optimal value of $head\_size$ was selected amongst 2, 4, 6, 8, 128 and 256.

From the above ranges, we obtained the optimal performance with $batch\_size$ = 64, $lr$ = 0.01, $gru\_units$ = 100, $dropout$ = 0.2, $num\_heads$ = 6, and $head\_size$ = 128.

\subsection{Experimental Results}

This section presents a detailed analysis of the model’s efficacy over our dataset. We compare the model’s performance with existing methods and measure the effectiveness of individual features on the model’s performance. We also present a qualitative analysis of the model’s performance over our dataset.

\subsubsection{Performance Comparison over the dataset}
To evaluate our proposed model's effectiveness, we compared its performance with existing methods. Each of the existing methods is implemented and trained over our dataset for prediction. Further, the models' performances are evaluated in terms of MAE and RMSE and reported in Table \ref{performancecomparision}. From the table, it can be observed that the model outperforms the existing methods while predicting new COVID-19 cases.

\begin{table}[h]
    \centering
    \begin{tabular}{lcc}
    \toprule
    Method                                                        & RMSE                & MAE\\
    \midrule
    Bedi \MakeLowercase{\textit{et al.}} LSTM~\cite{ref2}         & 52.49               & 27.88\\
    Arora \MakeLowercase{\textit{et al.}} LSTM~\cite{ref1}        & 61.66               & 37.11\\
    Arora \MakeLowercase{\textit{et al.}} Bi-LSTM~\cite{ref1}     & 73.50               & 56.97\\
    Arora \MakeLowercase{\textit{et al.}} Conv-LSTM~\cite{ref1}   & 89.60               & 61.06\\
    Shahid \MakeLowercase{\textit{et al.}} LSTM~\cite{ref3}       & 94.81               & 66.19\\
    Shahid \MakeLowercase{\textit{et al.}} Bi-LSTM~\cite{ref3}    & 95.06               & 65.76\\
    Shahid \MakeLowercase{\textit{et al.}} GRU~\cite{ref3}        & 94.23               & 67.25\\
    Proposed MAG model                                            & \textbf{22.81}      &  \textbf{11.35}\\
    \botrule
    \end{tabular}
    \caption{Performance Comparison on our dataset}
    \label{performancecomparision}
\end{table}

The proposed model shows an improvement of 29.68 and 16.53 over Bedi \MakeLowercase{\textit{et al.}}~\cite{ref2}, at least 38.85 and 25.76 over Arora \MakeLowercase{\textit{et al.}}~\cite{ref1}, and at least 71.42 and 54.84 over Shahid \MakeLowercase{\textit{et al.}} ~\cite{ref3}, in terms of RMSE and MAE. This performance improvement is the result of the multi-head attention block used in the MAG model. The attention block helps in capturing dependencies present between input data from different sources, that are, news headlines, google trends, and daily COVID statistics, and creates a rich joint representation, which in turn enhances the performance over basic time-series models.

\subsubsection{Performance Gain Ananlysis}
We measured the impact of each feature in predicting the number of COVID-19 cases. We first analyzed the impact of Google Trends on the model’s efficacy followed by News Encoding and different Statistical Combinations.

\begin{itemize}
    \item Google Trends - To measure the impact caused by the Google Trends data related to COVID-19 keywords, we evaluated our model by ruling out the trends feature while training the model and compared it with the performance of the original model. Table \ref{pcaGoogleTrends} presents the analysis results, where `With Google Trends' represents the original model and `Without Google Trends' represents the model trained without Google Trend features. It can be observed that there is an error reduction of 54.99 and 38.59 in terms of RMSE and MAE respectively. This can be attributed to the fact that the Google trends related to COVID-19 words align with the actual COVID-19 trend.

    \begin{table}[h]
        \caption{Performance Gain Analysis Result on Google Trends Feature}
        \centering
        \begin{tabular}{lcc}
        \toprule
        Method                            & RMSE               & MAE\\
        \midrule
        Without Google Trends             & 77.80              & 49.94\\
        With Google Trends                & \textbf{22.81}     & \textbf{11.35}\\
        \botrule
        \end{tabular}
        \label{pcaGoogleTrends}
    \end{table}

    \item News Encoding - We analyzed the impact of news encoding on the model’s performance by extracting every day’s news headlines related to COVID-19 and generating its encoding by passing through the BERT model. To measure the performance gain, we first evaluated our model by only feeding it the Google Trends and Statistical Features and comparing it with the results produced by the original model. Table \ref{pcaNewsEncoding} presents the results analyzing the performance gain due to news encoding. Without News Encoding represents the model trained without news encoding features. The original model shows an error reduction of 6.96 and 3.64 in terms of RMSE and MAE respectively, when compared to the model built by leaving the news encoding. This supports the claim that news headlines serve as an important source of information for forecasting COVID-19 cases.

    \begin{table}[h]
        \caption{Performance Gain Analysis Result on News Encoding Feature}
        \centering
        \begin{tabular}{lcc}
        \toprule
        Method                            & RMSE                & MAE\\
        \midrule
        Without News Encoding             & 29.77               & 14.99\\
        With News Encoding                & \textbf{22.81}      & \textbf{11.35}\\
        \botrule
        \end{tabular}
        \label{pcaNewsEncoding}
    \end{table}

    \item Statistical Combination - We present an analysis of the impact of individual statistical features on the model’s performance. To achieve this, we evaluated our proposed methodology by training it over different combinations of features. The combinations were built using a leave-one-out strategy, where we leave one of the features and feed the model with the remaining ones. The statistical features as mentioned in the proposed methodology and dataset description are (i) Cases, (ii) Vaccination, (iii) Test, (iv) Covid Patient, (v) Hospital, (vi) Population, and (vii) Other Countries.

    \begin{table}[h]
        \caption{Performance Gain Analysis Result on Statistical Combination}
        \centering
        \begin{tabular}{lcc}
        \toprule
        Method                                     & RMSE               & MAE\\
        \midrule
        All features except Cases                  & 61.05              & 49.46\\
        All features except Vaccinations           & 53.76              & 42.30\\
        All features except Test                   & 23.77              & 14.36\\
        All features except Covid Patient          & 74.97              & 47.20\\
        All features except Hospital               & 55.29              & 30.74\\
        All features except Population             & 59.01              & 46.35\\
        All features except Other Countries        & 48.74              & 29.79\\
        All features                               & \textbf{22.81}     & \textbf{11.35}\\
        \botrule
        \end{tabular}
        \label{pcaStatisticalCombinations}
    \end{table}
    
    Table \ref{pcaStatisticalCombinations} presents the results produced by the proposed model over the various combinations provided to it. For instance, `All features except Cases' represents the model trained on Vaccination, Test, Covid Patient, Hospital, Population, and Other Countries’ data while leaving out Cases (the number of new cases and deaths). The rest of the combinations are similarly represented in the table. It can be seen that the original model surpasses all the other combinations. This shows that each of the statistical features has a positive impact on the model’s performance.

    \item Multi-head Attention Block - In this section, we present an experimental analysis to evaluate the contribution of multi-head attention towards the model's performance. We removed the attention block from the proposed model and trained a basic two-layered GRU network over our dataset. Table \ref{pcaAttention} reports the RMSE and MAE produced by the model without attention block and compares the results with our proposed model. It can be seen that our proposed model surpasses the basic GRU model in terms of both RMSE and MAE. This is because the attention block can efficiently capture the dependency patterns among the input from different sources and produce an intermediate representation, which in turn enables improved performance over a basic GRU network.

    \begin{table}[h]
        \caption{Performance Gain Analysis Result on Multi-head Attention Block}
        \centering
        \begin{tabular}{lcc}
        \toprule
        Method                               & RMSE               & MAE\\
        \midrule
        Without Attention Block              & 53.72              & 24.79\\
        With Attention Block                 & \textbf{22.81}     & \textbf{11.35}\\
        \botrule
        \end{tabular}
        \label{pcaAttention}
    \end{table}
    
\end{itemize}

\subsubsection{Qualitative Analysis}

\begin{figure}[H]
\centering
\includegraphics[width=0.48\textwidth]{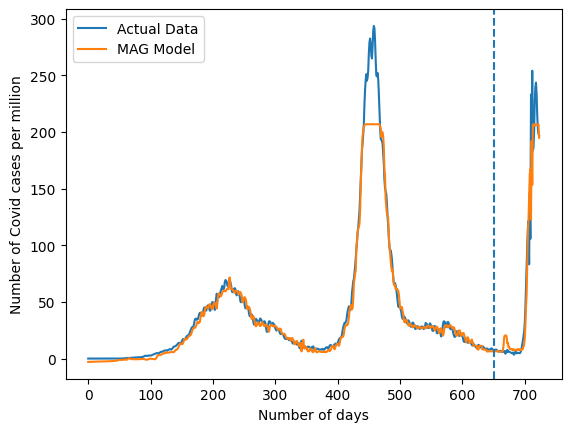}
\caption{Graph showing actual data vs predicted data}
\label{qualAnalysis}
\end{figure}

In this section, the model's qualitative performance is discussed. Fig \ref{qualAnalysis} shows a graphical analysis of the actual count of new COVID-19 cases vs the number of new cases predicted by the model 3 days prior. The vertical line splits the training data and the testing data. With a ratio of 90:10 between the training and testing set, the model gets trained over 653 days of data and predicts the number of cases for 73 days. It can be concluded from the figure that the model produces expected results and follows the daily trend of COVID cases, both on the training as well as testing datasets. While it can be seen that the model is not able to capture the peaks where the number of cases surpasses 200 cases per million, it perfectly captures the regular trends including the rise, decline, and shorter peaks.

\begin{figure}[H]
\centering
\includegraphics[width=0.48\textwidth]{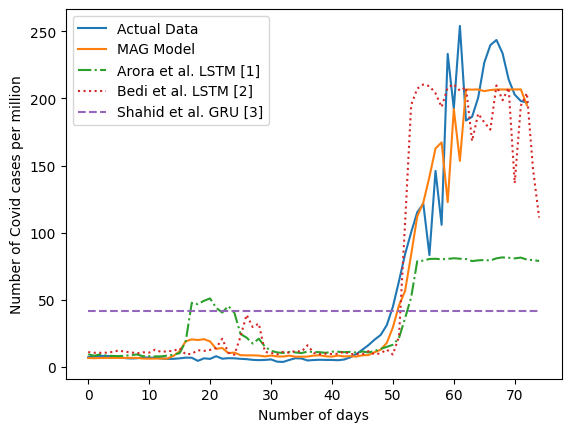}
\caption{Performance comparison graph over test data}
\label{qualAnalysisTest}
\end{figure}

Fig \ref{qualAnalysisTest} presents a zoomed-in view of the model’s performance on the testing dataset. It also shows the predictions made by the existing methods over the testing data. While each of the existing methods has implemented multiple models, we selected the model with the least RMSE for the graph. It can be seen that the MAG model clearly follows the trend better in comparison to the existing methodologies.

\begin{table}[h]
\caption{Similarity analysis of different method's prediction with actual data}
\centering
\begin{tabular}{lc}
\toprule
Method                                                        & Area between the curves  \\
\midrule
Arora \MakeLowercase{\textit{et al.}} LSTM~\cite{ref1}        & 2579.50           \\
Arora \MakeLowercase{\textit{et al.}} Bi-LSTM~\cite{ref1}     & 3701.29           \\
Arora \MakeLowercase{\textit{et al.}} Conv-LSTM~\cite{ref1}   & 3756.07           \\
Bedi \MakeLowercase{\textit{et al.}} LSTM~\cite{ref2}         & 1695.19           \\
Shahid \MakeLowercase{\textit{et al.}} LSTM~\cite{ref3}       & 4453.57           \\
Shahid \MakeLowercase{\textit{et al.}} Bi-LSTM~\cite{ref3}    & 4424.46           \\
Shahid \MakeLowercase{\textit{et al.}} GRU~\cite{ref3}        & 4525.46           \\
Proposed MAG Model                                            & \textbf{826.85}   \\
\botrule
\end{tabular}
\label{pearsoncoefficient}
\end{table}

Table \ref{pearsoncoefficient} shows the area between the curve generated by the individual model's prediction and the actual COVID-19 trajectory over the test dataset. A lower area shows that the two curves are closer and hence indicate better performance. To obtain the area between the two curves, we adopted the similarity measures\footnote{Similarity Measures - https://pypi.org/project/similaritymeasures/}\cite{ref37}. It can be seen that our model beats the existing methods with the lowest area of 826.85. It can be concluded that the model prediction is strongly related to and follows the same trend as the actual data.

\section{Conclusion}
\label{sec:sa_cnfw}
This paper aims to predict the number of cases during an infectious disease outbreak, three days in advance. To achieve this, we extract data from three sources, (i) News Headlines, (ii) Google Trends, and (iii) Statistical Data. Further, we propose a Multilateral Attention-enhanced GRU (MAG) model comprising (i) a Multi-head Attention block for comprehending inter-dependency in data from the three sources mentioned earlier and generating a rich joint representation, and (ii) two GRU layers for forecasting the output number of covid cases. The proposed model is trained and assessed using data collected for COVID-19 from $30^{th}$ January 2020 to $24^{th}$ January 2022. The experimental findings demonstrate that the proposed methodology surpasses the existing methods. Further analysis shows the impact of individual data sources and attention blocks on the model's performance. Finally, qualitative analysis shows that the proposed model closely follows the covid trend.

\section*{Declarations}

\begin{itemize}
\item \textbf{Funding} - The authors have no relevant financial or non-financial interests to disclose. 
\item \textbf{Competing Interests} - The authors have no conflicts of interest to declare that are relevant to the content of this article.
\item \textbf{Author contribution}
    \begin{itemize}
        \item Ashutosh Anshul: Conceptualization, Methodology, Writing- Original draft preparation, Data curation
        \item Jhalak Gupta: Data curation, Methodology, Investigation, Writing- Reviewing and Editing
        \item Mohammad Zia Ur Rehman: Methodology, Writing- Reviewing and Editing
        \item Nagendra Kumar: Supervision, Methodology, Writing- Reviewing and Editing
    \end{itemize}
\end{itemize}

\bibliography{sn-bibliography}


\begin{thebibliography}{40}
\ifx \bisbn   \undefined \def \bisbn  #1{ISBN #1}\fi
\ifx \binits  \undefined \def \binits#1{#1}\fi
\ifx \bauthor  \undefined \def \bauthor#1{#1}\fi
\ifx \batitle  \undefined \def \batitle#1{#1}\fi
\ifx \bjtitle  \undefined \def \bjtitle#1{#1}\fi
\ifx \bvolume  \undefined \def \bvolume#1{\textbf{#1}}\fi
\ifx \byear  \undefined \def \byear#1{#1}\fi
\ifx \bissue  \undefined \def \bissue#1{#1}\fi
\ifx \bfpage  \undefined \def \bfpage#1{#1}\fi
\ifx \blpage  \undefined \def \blpage #1{#1}\fi
\ifx \burl  \undefined \def \burl#1{\textsf{#1}}\fi
\ifx \doiurl  \undefined \def \doiurl#1{\url{https://doi.org/#1}}\fi
\ifx \betal  \undefined \def \betal{\textit{et al.}}\fi
\ifx \binstitute  \undefined \def \binstitute#1{#1}\fi
\ifx \binstitutionaled  \undefined \def \binstitutionaled#1{#1}\fi
\ifx \bctitle  \undefined \def \bctitle#1{#1}\fi
\ifx \beditor  \undefined \def \beditor#1{#1}\fi
\ifx \bpublisher  \undefined \def \bpublisher#1{#1}\fi
\ifx \bbtitle  \undefined \def \bbtitle#1{#1}\fi
\ifx \bedition  \undefined \def \bedition#1{#1}\fi
\ifx \bseriesno  \undefined \def \bseriesno#1{#1}\fi
\ifx \blocation  \undefined \def \blocation#1{#1}\fi
\ifx \bsertitle  \undefined \def \bsertitle#1{#1}\fi
\ifx \bsnm \undefined \def \bsnm#1{#1}\fi
\ifx \bsuffix \undefined \def \bsuffix#1{#1}\fi
\ifx \bparticle \undefined \def \bparticle#1{#1}\fi
\ifx \barticle \undefined \def \barticle#1{#1}\fi
\bibcommenthead
\ifx \bconfdate \undefined \def \bconfdate #1{#1}\fi
\ifx \botherref \undefined \def \botherref #1{#1}\fi
\ifx \url \undefined \def \url#1{\textsf{#1}}\fi
\ifx \bchapter \undefined \def \bchapter#1{#1}\fi
\ifx \bbook \undefined \def \bbook#1{#1}\fi
\ifx \bcomment \undefined \def \bcomment#1{#1}\fi
\ifx \oauthor \undefined \def \oauthor#1{#1}\fi
\ifx \citeauthoryear \undefined \def \citeauthoryear#1{#1}\fi
\ifx \endbibitem  \undefined \def \endbibitem {}\fi
\ifx \bconflocation  \undefined \def \bconflocation#1{#1}\fi
\ifx \arxivurl  \undefined \def \arxivurl#1{\textsf{#1}}\fi
\csname PreBibitemsHook\endcsname

\bibitem[\protect\citeauthoryear{Castillo and Melin}{2020}]{ref12}
\begin{barticle}
\bauthor{\bsnm{Castillo}, \binits{O.}},
\bauthor{\bsnm{Melin}, \binits{P.}}:
\batitle{Forecasting of {COVID}-19 time series for countries in the world based on a hybrid approach combining the fractal dimension and fuzzy logic}.
\bjtitle{Chaos, Solitons \& Fractals}
\bvolume{140},
\bfpage{110242}
(\byear{2020})
\doiurl{10.1016/j.chaos.2020.110242}
\end{barticle}
\endbibitem

\bibitem[\protect\citeauthoryear{Ma et~al.}{2021}]{ref34}
\begin{botherref}
\oauthor{\bsnm{Ma}, \binits{R.}},
\oauthor{\bsnm{Zheng}, \binits{X.}},
\oauthor{\bsnm{Wang}, \binits{P.}},
\oauthor{\bsnm{Liu}, \binits{H.}},
\oauthor{\bsnm{Zhang}, \binits{C.}}:
The prediction and analysis of {COVID}-19 epidemic trend by combining {LSTM} and markov method.
Scientific Reports
\textbf{11}(1)
(2021)
\doiurl{10.1038/s41598-021-97037-5}
\end{botherref}
\endbibitem

\bibitem[\protect\citeauthoryear{Bedi et~al.}{2021}]{ref2}
\begin{botherref}
\oauthor{\bsnm{Bedi}, \binits{P.}},
\oauthor{\bsnm{Dhiman}, \binits{S.}},
\oauthor{\bsnm{Gole}, \binits{P.}},
\oauthor{\bsnm{Gupta}, \binits{N.}},
\oauthor{\bsnm{Jindal}, \binits{V.}}:
Prediction of {COVID}-19 trend in india and its four worst-affected states using modified {SEIRD} and {LSTM} models.
{SN} Computer Science
\textbf{2}(3)
(2021)
\doiurl{10.1007/s42979-021-00598-5}
\end{botherref}
\endbibitem

\bibitem[\protect\citeauthoryear{Arora and Taylor}{2013}]{ref5}
\begin{barticle}
\bauthor{\bsnm{Arora}, \binits{S.}},
\bauthor{\bsnm{Taylor}, \binits{J.W.}}:
\batitle{Short-term forecasting of anomalous load using rule-based triple seasonal methods}.
\bjtitle{IEEE Transactions on Power Systems}
\bvolume{28}(\bissue{3}),
\bfpage{3235}--\blpage{3242}
(\byear{2013})
\doiurl{10.1109/TPWRS.2013.2252929}
\end{barticle}
\endbibitem

\bibitem[\protect\citeauthoryear{Maciel and Ballini}{2017}]{ref6}
\begin{bchapter}
\bauthor{\bsnm{Maciel}, \binits{L.}},
\bauthor{\bsnm{Ballini}, \binits{R.}}:
\bctitle{Interval fuzzy rule-based modeling approach for financial time series forecasting}.
In: \bbtitle{2017 IEEE International Conference on Fuzzy Systems (FUZZ-IEEE)},
pp. \bfpage{1}--\blpage{6}
(\byear{2017}).
\doiurl{10.1109/FUZZ-IEEE.2017.8015654}
\end{bchapter}
\endbibitem

\bibitem[\protect\citeauthoryear{Borisov and Luferov}{2019}]{ref7}
\begin{bchapter}
\bauthor{\bsnm{Borisov}, \binits{V.}},
\bauthor{\bsnm{Luferov}, \binits{V.}}:
\bctitle{Forecasting of multidimensional time series basing on fuzzy rule-based models}.
In: \bbtitle{2019 XXI International Conference Complex Systems: Control and Modeling Problems (CSCMP)},
pp. \bfpage{96}--\blpage{99}
(\byear{2019}).
\doiurl{10.1109/CSCMP45713.2019.8976821}
\end{bchapter}
\endbibitem

\bibitem[\protect\citeauthoryear{Rajagopalan and Rajamani}{2013}]{ref8}
\begin{bchapter}
\bauthor{\bsnm{Rajagopalan}, \binits{S.}},
\bauthor{\bsnm{Rajamani}, \binits{L.}}:
\bctitle{A fuzzy logic rule based forecasting model: Work-life balance in it among software vs. services industry on the view of women software engineer}.
In: \bbtitle{2013 International Conference on Machine Intelligence and Research Advancement},
pp. \bfpage{241}--\blpage{246}
(\byear{2013}).
\doiurl{10.1109/ICMIRA.2013.52}
\end{bchapter}
\endbibitem

\bibitem[\protect\citeauthoryear{Hakimi et~al.}{2021}]{ref9}
\begin{barticle}
\bauthor{\bsnm{Hakimi}, \binits{A.}},
\bauthor{\bsnm{Monadjemi}, \binits{S.A.}},
\bauthor{\bsnm{Setayeshi}, \binits{S.}}:
\batitle{An introduction of a reward-based time-series forecasting model and its application in predicting the dynamic and complicated behavior of the earth rotation (delta-t values)}.
\bjtitle{Applied Soft Computing}
\bvolume{113},
\bfpage{107920}
(\byear{2021})
\doiurl{10.1016/j.asoc.2021.107920}
\end{barticle}
\endbibitem

\bibitem[\protect\citeauthoryear{Arora and Taylor}{2018}]{ref10}
\begin{barticle}
\bauthor{\bsnm{Arora}, \binits{S.}},
\bauthor{\bsnm{Taylor}, \binits{J.W.}}:
\batitle{Rule-based autoregressive moving average models for forecasting load on special days: A case study for france}.
\bjtitle{European Journal of Operational Research}
\bvolume{266}(\bissue{1}),
\bfpage{259}--\blpage{268}
(\byear{2018})
\doiurl{10.1016/j.ejor.2017.08.056}
\end{barticle}
\endbibitem

\bibitem[\protect\citeauthoryear{Kamley et~al.}{2017}]{ref11}
\begin{bchapter}
\bauthor{\bsnm{Kamley}, \binits{S.}},
\bauthor{\bsnm{Jaloree}, \binits{S.}},
\bauthor{\bsnm{Thakur}, \binits{R.S.}}:
\bctitle{Forecasting of major world stock exchanges using rule-based forward and backward chaining expert systems}.
In: \bbtitle{Quality, {IT} and Business Operations},
pp. \bfpage{297}--\blpage{306}.
\bpublisher{Springer}, \blocation{???}
(\byear{2017}).
\doiurl{10.1007/978-981-10-5577-5_23}
\end{bchapter}
\endbibitem

\bibitem[\protect\citeauthoryear{Taylor}{2010}]{ref13}
\begin{barticle}
\bauthor{\bsnm{Taylor}, \binits{J.W.}}:
\batitle{Triple seasonal methods for short-term electricity demand forecasting}.
\bjtitle{European Journal of Operational Research}
\bvolume{204}(\bissue{1}),
\bfpage{139}--\blpage{152}
(\byear{2010})
\doiurl{10.1016/j.ejor.2009.10.003}
\end{barticle}
\endbibitem

\bibitem[\protect\citeauthoryear{Contreras et~al.}{2003}]{ref20}
\begin{barticle}
\bauthor{\bsnm{Contreras}, \binits{J.}},
\bauthor{\bsnm{Espinola}, \binits{R.}},
\bauthor{\bsnm{Nogales}, \binits{F.J.}},
\bauthor{\bsnm{Conejo}, \binits{A.J.}}:
\batitle{Arima models to predict next-day electricity prices}.
\bjtitle{IEEE Transactions on Power Systems}
\bvolume{18}(\bissue{3}),
\bfpage{1014}--\blpage{1020}
(\byear{2003})
\doiurl{10.1109/TPWRS.2002.804943}
\end{barticle}
\endbibitem

\bibitem[\protect\citeauthoryear{Ediger and Akar}{2007}]{ref21}
\begin{barticle}
\bauthor{\bsnm{Ediger}, \binits{V.{\c{S}}.}},
\bauthor{\bsnm{Akar}, \binits{S.}}:
\batitle{{ARIMA} forecasting of primary energy demand by fuel in turkey}.
\bjtitle{Energy Policy}
\bvolume{35}(\bissue{3}),
\bfpage{1701}--\blpage{1708}
(\byear{2007})
\doiurl{10.1016/j.enpol.2006.05.009}
\end{barticle}
\endbibitem

\bibitem[\protect\citeauthoryear{Guo et~al.}{2019}]{ref22}
\begin{barticle}
\bauthor{\bsnm{Guo}, \binits{J.}},
\bauthor{\bsnm{He}, \binits{H.}},
\bauthor{\bsnm{Sun}, \binits{C.}}:
\batitle{Arima-based road gradient and vehicle velocity prediction for hybrid electric vehicle energy management}.
\bjtitle{IEEE Transactions on Vehicular Technology}
\bvolume{68}(\bissue{6}),
\bfpage{5309}--\blpage{5320}
(\byear{2019})
\doiurl{10.1109/TVT.2019.2912893}
\end{barticle}
\endbibitem

\bibitem[\protect\citeauthoryear{Youness and Driss}{2022}]{ref18}
\begin{bchapter}
\bauthor{\bsnm{Youness}, \binits{J.}},
\bauthor{\bsnm{Driss}, \binits{M.}}:
\bctitle{An arima model for modeling and forecasting the dynamic of univariate time series: The case of moroccan inflation rate}.
In: \bbtitle{2022 International Conference on Intelligent Systems and Computer Vision (ISCV)},
pp. \bfpage{1}--\blpage{5}
(\byear{2022}).
\doiurl{10.1109/ISCV54655.2022.9806073}
\end{bchapter}
\endbibitem

\bibitem[\protect\citeauthoryear{Yang and Lin}{2016}]{ref23}
\begin{barticle}
\bauthor{\bsnm{Yang}, \binits{H.-L.}},
\bauthor{\bsnm{Lin}, \binits{H.-C.}}:
\batitle{An integrated model combined {ARIMA}, {EMD} with {SVR} for stock indices forecasting}.
\bjtitle{International Journal on Artificial Intelligence Tools}
\bvolume{25}(\bissue{02}),
\bfpage{1650005}
(\byear{2016})
\doiurl{10.1142/s0218213016500056}
\end{barticle}
\endbibitem

\bibitem[\protect\citeauthoryear{Li and Chiang}{2013}]{ref24}
\begin{barticle}
\bauthor{\bsnm{Li}, \binits{C.}},
\bauthor{\bsnm{Chiang}, \binits{T.-W.}}:
\batitle{Complex neurofuzzy arima forecasting—a new approach using complex fuzzy sets}.
\bjtitle{IEEE Transactions on Fuzzy Systems}
\bvolume{21}(\bissue{3}),
\bfpage{567}--\blpage{584}
(\byear{2013})
\doiurl{10.1109/TFUZZ.2012.2226890}
\end{barticle}
\endbibitem

\bibitem[\protect\citeauthoryear{Ord{\'{o}}{\~{n}}ez et~al.}{2019}]{ref25}
\begin{barticle}
\bauthor{\bsnm{Ord{\'{o}}{\~{n}}ez}, \binits{C.}},
\bauthor{\bsnm{Lasheras}, \binits{F.S.}},
\bauthor{\bsnm{Roca-Pardi{\~{n}}as}, \binits{J.}},
\bauthor{\bsnm{Cos~Juez}, \binits{F.J.}}:
\batitle{A hybrid {ARIMA}{\textendash}{SVM} model for the study of the remaining useful life of aircraft engines}.
\bjtitle{Journal of Computational and Applied Mathematics}
\bvolume{346},
\bfpage{184}--\blpage{191}
(\byear{2019})
\doiurl{10.1016/j.cam.2018.07.008}
\end{barticle}
\endbibitem

\bibitem[\protect\citeauthoryear{Sharma et~al.}{2021}]{ref15}
\begin{barticle}
\bauthor{\bsnm{Sharma}, \binits{R.R.}},
\bauthor{\bsnm{Kumar}, \binits{M.}},
\bauthor{\bsnm{Maheshwari}, \binits{S.}},
\bauthor{\bsnm{Ray}, \binits{K.P.}}:
\batitle{Evdhm-arima-based time series forecasting model and its application for covid-19 cases}.
\bjtitle{IEEE Transactions on Instrumentation and Measurement}
\bvolume{70},
\bfpage{1}--\blpage{10}
(\byear{2021})
\doiurl{10.1109/TIM.2020.3041833}
\end{barticle}
\endbibitem

\bibitem[\protect\citeauthoryear{Bang et~al.}{2019}]{ref26}
\begin{bchapter}
\bauthor{\bsnm{Bang}, \binits{S.}},
\bauthor{\bsnm{Bishnoi}, \binits{R.}},
\bauthor{\bsnm{Chauhan}, \binits{A.S.}},
\bauthor{\bsnm{Dixit}, \binits{A.K.}},
\bauthor{\bsnm{Chawla}, \binits{I.}}:
\bctitle{Fuzzy logic based crop yield prediction using temperature and rainfall parameters predicted through arma, sarima, and armax models}.
In: \bbtitle{2019 Twelfth International Conference on Contemporary Computing (IC3)},
pp. \bfpage{1}--\blpage{6}
(\byear{2019}).
\doiurl{10.1109/IC3.2019.8844901}
\end{bchapter}
\endbibitem

\bibitem[\protect\citeauthoryear{Chanpanit et~al.}{2019}]{ref27}
\begin{bchapter}
\bauthor{\bsnm{Chanpanit}, \binits{T.}},
\bauthor{\bsnm{Arkamanont}, \binits{N.}},
\bauthor{\bsnm{Pranootnarapran}, \binits{N.}}:
\bctitle{Predicting the number of people for road traffic accident on highways by hour of day}.
In: \bbtitle{2019 8th International Conference on Industrial Technology and Management (ICITM)},
pp. \bfpage{285}--\blpage{289}
(\byear{2019}).
\doiurl{10.1109/ICITM.2019.8710718}
\end{bchapter}
\endbibitem

\bibitem[\protect\citeauthoryear{Dwivedi et~al.}{2021}]{ref16}
\begin{bchapter}
\bauthor{\bsnm{Dwivedi}, \binits{S.A.}},
\bauthor{\bsnm{Attry}, \binits{A.}},
\bauthor{\bsnm{Parekh}, \binits{D.}},
\bauthor{\bsnm{Singla}, \binits{K.}}:
\bctitle{Analysis and forecasting of time-series data using s-arima, cnn and lstm}.
In: \bbtitle{2021 International Conference on Computing, Communication, and Intelligent Systems (ICCCIS)},
pp. \bfpage{131}--\blpage{136}
(\byear{2021}).
\doiurl{10.1109/ICCCIS51004.2021.9397134}
\end{bchapter}
\endbibitem

\bibitem[\protect\citeauthoryear{Siami-Namini et~al.}{2018}]{ref14}
\begin{bchapter}
\bauthor{\bsnm{Siami-Namini}, \binits{S.}},
\bauthor{\bsnm{Tavakoli}, \binits{N.}},
\bauthor{\bsnm{Siami~Namin}, \binits{A.}}:
\bctitle{A comparison of arima and lstm in forecasting time series}.
In: \bbtitle{2018 17th IEEE International Conference on Machine Learning and Applications (ICMLA)},
pp. \bfpage{1394}--\blpage{1401}
(\byear{2018}).
\doiurl{10.1109/ICMLA.2018.00227}
\end{bchapter}
\endbibitem

\bibitem[\protect\citeauthoryear{Sirisha et~al.}{2022}]{ref17}
\begin{barticle}
\bauthor{\bsnm{Sirisha}, \binits{U.M.}},
\bauthor{\bsnm{Belavagi}, \binits{M.C.}},
\bauthor{\bsnm{Attigeri}, \binits{G.}}:
\batitle{Profit prediction using arima, sarima and lstm models in time series forecasting: A comparison}.
\bjtitle{IEEE Access}
\bvolume{10},
\bfpage{124715}--\blpage{124727}
(\byear{2022})
\doiurl{10.1109/ACCESS.2022.3224938}
\end{barticle}
\endbibitem

\bibitem[\protect\citeauthoryear{Ghanbari and Borna}{2021}]{ref28}
\begin{bchapter}
\bauthor{\bsnm{Ghanbari}, \binits{R.}},
\bauthor{\bsnm{Borna}, \binits{K.}}:
\bctitle{Multivariate time-series prediction using lstm neural networks}.
In: \bbtitle{2021 26th International Computer Conference, Computer Society of Iran (CSICC)},
pp. \bfpage{1}--\blpage{5}
(\byear{2021}).
\doiurl{10.1109/CSICC52343.2021.9420543}
\end{bchapter}
\endbibitem

\bibitem[\protect\citeauthoryear{Wang et~al.}{2019}]{ref29}
\begin{bchapter}
\bauthor{\bsnm{Wang}, \binits{Y.}},
\bauthor{\bsnm{Zhu}, \binits{S.}},
\bauthor{\bsnm{Li}, \binits{C.}}:
\bctitle{Research on multistep time series prediction based on lstm}.
In: \bbtitle{2019 3rd International Conference on Electronic Information Technology and Computer Engineering (EITCE)},
pp. \bfpage{1155}--\blpage{1159}
(\byear{2019}).
\doiurl{10.1109/EITCE47263.2019.9095044}
\end{bchapter}
\endbibitem

\bibitem[\protect\citeauthoryear{Zhai et~al.}{2020}]{ref30}
\begin{bchapter}
\bauthor{\bsnm{Zhai}, \binits{N.}},
\bauthor{\bsnm{Yao}, \binits{P.}},
\bauthor{\bsnm{Zhou}, \binits{X.}}:
\bctitle{Multivariate time series forecast in industrial process based on xgboost and gru}.
In: \bbtitle{2020 IEEE 9th Joint International Information Technology and Artificial Intelligence Conference (ITAIC)},
vol. \bseriesno{9},
pp. \bfpage{1397}--\blpage{1400}
(\byear{2020}).
\doiurl{10.1109/ITAIC49862.2020.9338878}
\end{bchapter}
\endbibitem

\bibitem[\protect\citeauthoryear{Honjo et~al.}{2022}]{ref31}
\begin{bchapter}
\bauthor{\bsnm{Honjo}, \binits{K.}},
\bauthor{\bsnm{Zhou}, \binits{X.}},
\bauthor{\bsnm{Shimizu}, \binits{S.}}:
\bctitle{Cnn-gru based deep learning model for demand forecast in retail industry}.
In: \bbtitle{2022 International Joint Conference on Neural Networks (IJCNN)},
pp. \bfpage{1}--\blpage{8}
(\byear{2022}).
\doiurl{10.1109/IJCNN55064.2022.9892599}
\end{bchapter}
\endbibitem

\bibitem[\protect\citeauthoryear{Xie et~al.}{2017}]{ref32}
\begin{bchapter}
\bauthor{\bsnm{Xie}, \binits{R.}},
\bauthor{\bsnm{Ding}, \binits{Y.}},
\bauthor{\bsnm{Hao}, \binits{K.}},
\bauthor{\bsnm{Chen}, \binits{L.}},
\bauthor{\bsnm{Wang}, \binits{T.}}:
\bctitle{Using gated recurrence units neural network for prediction of melt spinning properties}.
In: \bbtitle{2017 11th Asian Control Conference (ASCC)},
pp. \bfpage{2286}--\blpage{2291}
(\byear{2017}).
\doiurl{10.1109/ASCC.2017.8287531}
\end{bchapter}
\endbibitem

\bibitem[\protect\citeauthoryear{Wang et~al.}{2022}]{ref33}
\begin{botherref}
\oauthor{\bsnm{Wang}, \binits{S.}},
\oauthor{\bsnm{Shao}, \binits{C.}},
\oauthor{\bsnm{Zhang}, \binits{J.}},
\oauthor{\bsnm{Zheng}, \binits{Y.}},
\oauthor{\bsnm{Meng}, \binits{M.}}:
Traffic flow prediction using bi-directional gated recurrent unit method.
Urban Informatics
\textbf{1}(1)
(2022)
\doiurl{10.1007/s44212-022-00015-z}
\end{botherref}
\endbibitem

\bibitem[\protect\citeauthoryear{Arora et~al.}{2020}]{ref1}
\begin{barticle}
\bauthor{\bsnm{Arora}, \binits{P.}},
\bauthor{\bsnm{Kumar}, \binits{H.}},
\bauthor{\bsnm{Panigrahi}, \binits{B.K.}}:
\batitle{Prediction and analysis of covid-19 positive cases using deep learning models: A descriptive case study of india}.
\bjtitle{Chaos, Solitons \& Fractals}
\bvolume{139},
\bfpage{110017}
(\byear{2020})
\doiurl{10.1016/j.chaos.2020.110017}
\end{barticle}
\endbibitem

\bibitem[\protect\citeauthoryear{Shahid et~al.}{2020}]{ref3}
\begin{barticle}
\bauthor{\bsnm{Shahid}, \binits{F.}},
\bauthor{\bsnm{Zameer}, \binits{A.}},
\bauthor{\bsnm{Muneeb}, \binits{M.}}:
\batitle{Predictions for {COVID}-19 with deep learning models of {LSTM}, {GRU} and bi-{LSTM}}.
\bjtitle{Chaos, Solitons \& Fractals}
\bvolume{140},
\bfpage{110212}
(\byear{2020})
\doiurl{10.1016/j.chaos.2020.110212}
\end{barticle}
\endbibitem

\bibitem[\protect\citeauthoryear{Jain et~al.}{2021}]{ref19}
\begin{bchapter}
\bauthor{\bsnm{Jain}, \binits{A.}},
\bauthor{\bsnm{Sukhdeve}, \binits{T.}},
\bauthor{\bsnm{Gadia}, \binits{H.}},
\bauthor{\bsnm{Sahu}, \binits{S.P.}},
\bauthor{\bsnm{Verma}, \binits{S.}}:
\bctitle{Covid19 prediction using time series analysis}.
In: \bbtitle{2021 International Conference on Artificial Intelligence and Smart Systems (ICAIS)},
pp. \bfpage{1599}--\blpage{1606}
(\byear{2021}).
\doiurl{10.1109/ICAIS50930.2021.9395877}
\end{bchapter}
\endbibitem

\bibitem[\protect\citeauthoryear{Alabdulrazzaq et~al.}{2021}]{ref35}
\begin{barticle}
\bauthor{\bsnm{Alabdulrazzaq}, \binits{H.}},
\bauthor{\bsnm{Alenezi}, \binits{M.N.}},
\bauthor{\bsnm{Rawajfih}, \binits{Y.}},
\bauthor{\bsnm{Alghannam}, \binits{B.A.}},
\bauthor{\bsnm{Al-Hassan}, \binits{A.A.}},
\bauthor{\bsnm{Al-Anzi}, \binits{F.S.}}:
\batitle{On the accuracy of {ARIMA} based prediction of {COVID}-19 spread}.
\bjtitle{Results in Physics}
\bvolume{27},
\bfpage{104509}
(\byear{2021})
\doiurl{10.1016/j.rinp.2021.104509}
\end{barticle}
\endbibitem

\bibitem[\protect\citeauthoryear{Rafiq et~al.}{2020}]{ref36}
\begin{barticle}
\bauthor{\bsnm{Rafiq}, \binits{D.}},
\bauthor{\bsnm{Suhail}, \binits{S.A.}},
\bauthor{\bsnm{Bazaz}, \binits{M.A.}}:
\batitle{Evaluation and prediction of {COVID}-19 in india: A case study of worst hit states}.
\bjtitle{Chaos, Solitons \& Fractals}
\bvolume{139},
\bfpage{110014}
(\byear{2020})
\doiurl{10.1016/j.chaos.2020.110014}
\end{barticle}
\endbibitem

\bibitem[\protect\citeauthoryear{K\"{u}lah et~al.}{2023}]{ref39}
\begin{barticle}
\bauthor{\bsnm{K\"{u}lah}, \binits{E.}},
\bauthor{\bsnm{{\c{C}}etinkaya}, \binits{Y.M.}},
\bauthor{\bsnm{\"{O}zer}, \binits{A.G.}},
\bauthor{\bsnm{Alemdar}, \binits{H.}}:
\batitle{{COVID}-19 forecasting using shifted gaussian mixture model with similarity-based estimation}.
\bjtitle{Expert Systems with Applications}
\bvolume{214},
\bfpage{119034}
(\byear{2023})
\doiurl{10.1016/j.eswa.2022.119034}
\end{barticle}
\endbibitem

\bibitem[\protect\citeauthoryear{Qu et~al.}{2023}]{ref38}
\begin{barticle}
\bauthor{\bsnm{Qu}, \binits{Z.}},
\bauthor{\bsnm{Li}, \binits{Y.}},
\bauthor{\bsnm{Jiang}, \binits{X.}},
\bauthor{\bsnm{Niu}, \binits{C.}}:
\batitle{An innovative ensemble model based on multiple neural networks and a novel heuristic optimization algorithm for {COVID}-19 forecasting}.
\bjtitle{Expert Systems with Applications}
\bvolume{212},
\bfpage{118746}
(\byear{2023})
\doiurl{10.1016/j.eswa.2022.118746}
\end{barticle}
\endbibitem

\bibitem[\protect\citeauthoryear{Ali et~al.}{2023}]{ref40}
\begin{barticle}
\bauthor{\bsnm{Ali}, \binits{F.}},
\bauthor{\bsnm{Ullah}, \binits{F.}},
\bauthor{\bsnm{Khan}, \binits{J.I.}},
\bauthor{\bsnm{Khan}, \binits{J.}},
\bauthor{\bsnm{Sardar}, \binits{A.W.}},
\bauthor{\bsnm{Lee}, \binits{S.}}:
\batitle{{COVID}-19 spread control policies based early dynamics forecasting using deep learning algorithm}.
\bjtitle{Chaos, Solitons \& Fractals}
\bvolume{167},
\bfpage{112984}
(\byear{2023})
\doiurl{10.1016/j.chaos.2022.112984}
\end{barticle}
\endbibitem

\bibitem[\protect\citeauthoryear{Devlin et~al.}{2019}]{ref4}
\begin{bchapter}
\bauthor{\bsnm{Devlin}, \binits{J.}},
\bauthor{\bsnm{Chang}, \binits{M.-W.}},
\bauthor{\bsnm{Lee}, \binits{K.}},
\bauthor{\bsnm{Toutanova}, \binits{K.}}:
\bctitle{{BERT}: Pre-training of deep bidirectional transformers for language understanding}.
In: \bbtitle{Proceedings of the 2019 Conference of the North {A}merican Chapter of the Association for Computational Linguistics: Human Language Technologies, Volume 1 (Long and Short Papers)},
pp. \bfpage{4171}--\blpage{4186}.
\bpublisher{Association for Computational Linguistics},
\blocation{Minneapolis, Minnesota}
(\byear{2019}).
\doiurl{10.18653/v1/N19-1423}
\end{bchapter}
\endbibitem

\bibitem[\protect\citeauthoryear{Jekel et~al.}{2018}]{ref37}
\begin{barticle}
\bauthor{\bsnm{Jekel}, \binits{C.F.}},
\bauthor{\bsnm{Venter}, \binits{G.}},
\bauthor{\bsnm{Venter}, \binits{M.P.}},
\bauthor{\bsnm{Stander}, \binits{N.}},
\bauthor{\bsnm{Haftka}, \binits{R.T.}}:
\batitle{Similarity measures for identifying material parameters from hysteresis loops using inverse analysis}.
\bjtitle{International Journal of Material Forming}
\bvolume{12}(\bissue{3}),
\bfpage{355}--\blpage{378}
(\byear{2018})
\doiurl{10.1007/s12289-018-1421-8}
\end{barticle}
\endbibitem

\end{thebibliography}

\end{document}